%% file: main.tex
\documentclass[11pt,twocolumn]{article}
\usepackage{graphicx}
\usepackage[utf8]{inputenc}
\usepackage[hidelinks]{hyperref}
\usepackage{pdfpages}
\usepackage{fancyhdr} 
\usepackage{lastpage}

\title{Fake News Detection using Stance Classification: A Survey}
\author{Anders E. Lillie (aedl@itu.dk) and Emil R. Middelboe (erem@itu.dk)}
\date{December 11, 2018}

\begin{document}
\maketitle
\thispagestyle{fancy}
\begin{abstract}
This paper surveys and presents recent academic work carried out within the field of stance classification and fake news detection. Echo chambers and the model organism problem are examples that pose challenges to acquire data with high quality, due to opinions being polarised in microblogs. Nevertheless it is shown that several machine learning approaches achieve promising results in classifying stance. Some use \textit{crowd stance} for fake news detection, such as the approach in \cite{dungs18} using Hidden Markov Models. Furthermore feature engineering have significant importance in several approaches, which is shown in \cite{aker17}. This paper additionally includes a proposal of a system implementation based on the presented survey. 
\end{abstract}
\input{introduction.tex}
\input{stance_and_fakenews.tex}
\input{data.tex}
\input{approaches.tex}
\input{proposal.tex}
\input{conclusion.tex}

\bibliographystyle{apalike}
\bibliography{articles}

\appendix
\input{appendix.tex}

\end{document}

%% file: introduction.tex
\section{Introduction}
\label{introduction}
Fake news detection currently relies on knowing the attitude that people communicating on social media are expressing towards an idea. Figuring this out is called stance classification, which is a Natural Language Processing (NLP) task that seeks to classify the stance taken towards some claim. This paper reviews different ideas and approaches towards accomplishing this goal. 

NLP is a research area concerned with processing human language using language models and computational approaches like machine learning (ML). With the progress of ML, tools and techniques open up for various ways of designing the algorithm for stance classification. It is interesting to investigate this progress and gain insight into current state-of-the-art approaches.

The work presented in this paper is carried out in the "Thesis Preparation" course at the IT-University of Copenhagen on the third semester of the MSc Software Development program. As such it is a project preparing for the thesis in Spring, 2019. The following is the tentative research question for the thesis project.

\subsection{Research question}
Stance classification and fake news detection is currently mostly concerned with the English language. The thesis project will attempt to answer the following questions: how do we build an automatic stance classification system for Danish? Further, how do we apply this system to verify or refute rumours and possibly detect fake news? 

\subsection{Overview}
The objective of this paper will thus be to study the approaches used for stance classification and fake news detection in the English language and what methods might be applicable to build a system for the Danish language. In particular section \ref{stance_classification} will provide context and definition for the term \textit{stance classification}. Section \ref{fake_news} will discuss definitions of fake news detection, refer to recent work and discuss a number of social and psychological aspects in the area. Section \ref{data} will cover data gathering, feature extraction and data annotation, as well as give context for the structure of microblogs. Section \ref{approaches} covers a number of different approaches taken to classify stance and detect fake news. Section \ref{proposal} will present proposals for the choice of approach, data gathering and technology for the thesis project, in addition to a high-level thesis plan. Finally section \ref{conclusion} will summarise the findings of this research paper.



%% file: stance_and_fakenews.tex
\section{Stance classification}
\label{stance_classification}
Literature on stance classification and stance detection systems is rather new, as most of the papers are published within the last 10 years. One of the first studies in the area is from \cite{qazvinian11}, in which they gather data from Twitter containing more than 10,000 tweets over 5 different topics. They propose a system for identifying misinformation in microblogs using different Bayes classifiers, and extracting ``content-based", ``network-based", and ``Twitter specific memes" features. Different approaches and objectives have since been set to tackle the computational task of classifying stance given some data based on a number of claims. 

Conversations in microblogs, such as Twitter, are typically used in classifying the stance for each reply to the source post, which expresses some claim. Many systems use the Support, Denying, Querying, and Commenting (SDQC) labels for classifying these posts\cite{zubiaga16}. Before stance classification is further investigated, we discuss applications of stance classification as well as related subjects. \\

\subsection{Applications}
\textit{Stance classification} is an area with closely related subjects, including \textit{veracity classification/detection} and \textit{fake news detection}. The reason for this is that stance classification can be used in the task of veracity classification, as well as fake news detection\cite{dungs18, shu2017fake}. In this paper the term \textit{stance classification} refers to the task of determining the opinion behind some text towards a specific target. As such, stance \textit{detection} is the task of using the classification system to automatically discover stance, and this term is used interchangeably with stance \textit{classification}. The same goes for \textit{veracity classification} which, on the other hand, is the task of resolving some claim by analysing crowd reactions\cite{derczynski17}. 

The task of stance classification often comes in two variants: \textit{open} and \textit{target-specific}\cite{aker17}. Open stance classification is applied in contexts, where no target/topic is known in advance, which makes it suitable for rumour resolution. Since the attitudes(stances) from a crowd towards some claim can be indicative of its truthfulness, it is as such applicable in veracity detection\cite{dungs18}.  In target-specific stance classification, on the other hand, cues about a target that is known in advance are provided in the training data. This can make classification of stance from unseen data, but with the same target, easier\cite{mohammad16}.

Furthermore the above described variants of stance classification can be either \textit{supervised} or \textit{unsupervised}. In the former case classification has prior knowledge based on a \textit{ground truth}, i.e. data is annotated, and in the latter case classification must be inferred from data, since there is no prior knowledge\footnote{\url{https://towardsdatascience.com/supervised-vs-unsupervised-learning-14f68e32ea8d}. Visited 03-12-2018}. \\

In the next section we introduce fake news detection and explore how stance classification is used for rumour resolution. 

\section{Fake news detection}
\label{fake_news}
One definition of fake news is that ``fake news is news articles that are intentionally and verifiably false"\cite{shu2017fake}. The key features of this statement is (1) authenticity: fake news include false information that can be verified, and (2) intent: fake news is created with dishonest intention to mislead consumers. A related area is that of rumour classification, in which the veracity of circulating information is yet to be verified at the time of spreading\cite{shu2017fake}. Thus the distinction is that fake news is intentionally misleading and is something which can be proven to be fake. The problem to solve for detecting rumours and fake news is however much the same. In the context of Twitter for example, given a source tweet containing a claim and a number of responses, the task is to determine whether the claim is true or false. 

\texttt{PHEME} is a project dealing with the fake news detection problem described above, focusing on veracity of data in social media and on the web\cite{pheme}. In particular four kinds of false claims are sought to be identified in real time: rumours, disinformation, misinformation, and speculation. Out of these four categories \textit{disinformation} most precisely describes the definition of fake news given above, i.e. information that is spread deliberately to deceive, in contrast to \textit{misinformation}, which is unintentional. Since the start of \texttt{PHEME} in 2014, several studies and papers have been published dealing with the task mentioned here, including \cite{kochlina17,derczynski17,zubiaga16}. 

The task of identifying false claims is also undertaken in the Fake News Challenge\cite{fakenewschallenge}. The goal in this challenge is to explore how ML and NLP can be used to combat the ``fake news problem". Specifically the task is broken down into stages, with the first stage being stance detection, classifying whether a body text agrees, disagrees, discusses or is unrelated to a headline. Note that this is quite different from the analysis of microblog data, where the posts in a sense are dynamic due to its temporal feature. However, related to the task of the Fake News Challenge is the work of \cite{augenstein16}, in which they build a classification system to interpret tweet stance towards previously unseen targets and where the target is not always mentioned in the text. Specifically they build a model to classify tweets regarding Donald Trump, where the training and development data is based on the targets \textit{Climate Change is a Real Concern}, \textit{Feminist Movement}, \textit{Atheism}, \textit{Legalization of Abortion}, and \textit{Hillary Clinton}.\\




\subsection{Social and psychological aspects}
\label{social_and_psycho}
Since fake news revolve around people it is interesting to investigate which social and psychological factors that have relevance and implications for fake news detection.

Some concepts that may have effects for the data used in fake news detection are \textit{confirmation bias} and \textit{the echo chamber effect}\cite{shu2017fake}. Confirmation bias describes consumers who prefer to receive information that confirms their existing views, while the echo chamber effect describes users on social media that tend to form groups containing like-minded people with polarised opinions. These phenomena are discussed in \cite{echochamber}, which carries out research on a large Facebook dataset. The research shows that users tend to polarise their interactions with users and pages of the same kind. Furthermore it is shown that the degree of polarisation correlates with the degree of sentiment extremity in the users' comments. \\

Another concept describing sharing of information between users is \textit{filter bubbles} and is covered in \cite{filter_bubbles}. Filter bubbles describe isolated users receiving news and information which does not overlap with information other users get. As such filter bubbles are much alike echo chambers, however \cite{filter_bubbles} has a focus on filter bubbles in relation to the Facebook news feed. The paper concludes that respectively 10.0 and 27.8 percentage of users in the used data set were in a filter bubble, depending on the approach. Furthermore it is noted that there is no clear connection between age, education, living location or gender and being in a filter bubble. However the users in filter bubbles had fewer friends, group likes and page likes than users who were not.\\

While \cite{filter_bubbles} and \cite{echochamber} both examine spread and isolation of information, it is important to note a key difference between them. \cite{filter_bubbles} covers information spread of news content specifically on the Facebook news feed in relation to the algorithm Edge Rank\footnote{\url{http://edgerank.net/} visited 09-12-2018}, while \cite{echochamber} examine the spread of information in regards to shared posts, page likes and so forth. 

The above findings show that it is important to keep these social and psychological aspects in mind, while considering the data used from social media platforms. Otherwise polarised or skewed data could have implications for the results and later usefulness of research in other contexts. This leads to the next section, where data and factors which influence its quality is discussed.

%% file: data.tex
\section{Data}
\label{data}
Gathering data for stance classification is a task in itself, as different factors, such as bias and class distribution, can have significant consequences for the resulting system. Social and psychological aspects in this regard are discussed above in section \ref{social_and_psycho}. Furthermore classifiers performs better with datasets with balanced class labels after annotation has been performed. Otherwise you might end up with misleading/imprecise classification systems: In \cite{kochlina17} they build the best-performing system for SemEval 2017 Task 8, subtask A, but due to unbalanced data, the model is primarily able to just classify ``commenting" instances, with only few correct predictions of ``denying" and ``supporting" instances, which are the more interesting classes.

\subsection{Data gathering}
This section will provide an overview of approaches to gather relevant data for the stance classification task.

In \cite{castillo11} a system is generated to gather data from Twitter and filter newsworthy topics. First they monitor Twitter posts in a period of 2 months using a monitoring system\footnote{``Twitter Monitor"(currently unavailable) from: \newline \url{http://www.twittermonitor.net/}}, which detects bursts in the frequency of sets of keywords found in messages. Then they query the system with specific keywords and collect tweets that match them during the burst peaks. They gather Twitter data in this way on over 2500 topics, and filter newsworthy ones from pure conversations with the MTurk API\footnote{\url{https://www.mturk.com/}}. The paper also describes how the labels given from MTurk is used to train a J48 decision tree classifier to filter the topics automatically.

Similarly a dataset is generated from Twitter using regular expression queries in \cite{qazvinian11}. They utilise the Twitter API by searching for data with queries that each represent a popular rumour that is deemed either ``false" or ``partly true" by About.com\footnote{\url{http://urbanlegends.about.com}}. Then two annotators manually go over all the tweets collected and annotate whether they are about a set of particular rumours.

More recent datasets include those in the SemEval tasks, such as SemEval 2016, task 6\cite{mohammad16} and SemEval 2017, Task 8\cite{derczynski17}. Alternative datasets are discussed in \cite{shu2017fake}, including BuzzFeedNews, LIAR, BS Detector, and CREDBANK. They point out, however, that these datasets have limitations that make them challenging for fake news detection. As a result they are currently in the process of developing their own dataset, which include news content and social context features\cite{shu18fakenewsnet}, which are the feature categories they find important for the fake news detection task. 


\subsection{Feature extraction}
Once data has been extracted for analysis, one must extract features relevant for the task at hand. The subject of \textit{feature engineering} could comprise a whole paper in itself. As such this section will not try to compare features, but will provide an overview of the most common features used for stance classification and fake news detection\cite{castillo11,shu2017fake,qazvinian11,aker17,kochlina17,enayet17}. Table \ref{tab:features} is a compact list of groups of similar features with accompanying short descriptions. Additionally a popular approach is to also include word embeddings using the \texttt{word2vec} algorithm\cite{word2vec}, representing words by dense vectors, as done in \cite{kochlina17}. \\

\begin{table}[h]
    \centering
    \begin{tabular}{l p{5cm}}
        \textit{Feature} &  \textit{Description} \\
        \hline
        Lexical & Count of words and characters, ratio of capital letters, names, as well as presence of period, question mark, exclamation mark, and special words (e.g. negation words) \\
        Attachments & URLs, images, and/or hashtags content \\
        Syntax & Sentence-level features, e.g. n-grams, BOW, POS tags \\
        User & No. of posts written, user creation date, no. of followers, demographics \\
        Post & Source or reply, relation to other posts, sentiment (positive/negative polarity), temporal information \\
    \end{tabular}
    \caption{An overview of the most common features used in stance classification and fake news detection}
    \label{tab:features}
\end{table}

With the progress of ML, tools and techniques open up for various ways of tackling the task of stance classification. Several studies however show that the most crucial part of stance classification is to extract and use the most \textit{optimal} features\cite{aker17,dungs18}. In early work, it was explored how features could be categorised into four classes, message-, topic-, user-, and propagation-based \cite{castillo11}. Although they are Twitter-specific, they are claimed to be generic. The message-based features deals with characteristics of messages, such as tweet-length. The user-based features, on the other hand, deals with characteristics of the user, which posts the message, such as registration age. Topic-based features are then an aggregation computed from the message- and user-based, such as the fraction of tweets containing URLs. Finally the propagation-based features consider the conversation tree, e.g. depth of the tree.

Another study shows that more or less abandoning the idea of having many features can provide significant results \cite{dungs18}. Their contribution shows how stance and tweet times alone achieve state-of-the-art results in the task of veracity detection as opposed to approaches using content and user based features as those introduced above. Along these lines \cite{aker17} shows how, by adding just six tweet confidence ``problem-specific" features to existing well-performing features, they achieve better results than previous systems on the same data. They prove this by using a decision tree stance classifier, which allegedly is simpler in its approach in comparison to competing systems'. 

\subsection{Data structure in microblogs}
\label{data_structure}
\begin{figure}[h]
    \centering
    \includegraphics[scale=0.9]{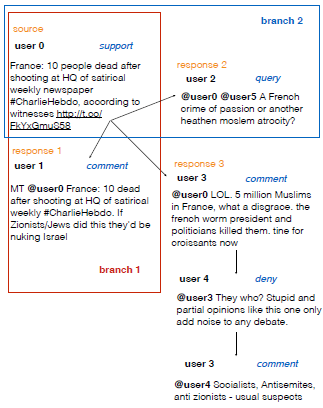}
    \caption{A conversation thread with three branches. Source: \cite{kochlina17}}
    \label{fig:conversation}
\end{figure}

This paper investigates stance classification over social media data and in particular from microblog platforms, as the structure makes it applicable for this task\cite{tolmie18}. As an example of a microblog conversation, figure \ref{fig:conversation} from \cite{kochlina17} illustrates a Twitter conversation, where a source post makes a claim and nested replies respond to it either directly or indirectly. Note that the tweets are also annotated, which is discussed in the next section. 

In \cite{procter13} they analyse how rumours propagate in Twitter, which we hypothesize also applies for similar microblogs such as Reddit\footnote{\url{https://www.reddit.com/}}. In short it comprises the following events, which we have reformulated to be general for microblogs:

\begin{enumerate}
    \item A rumour starts with someone posting about the occurrence of an alleged incident.
    \item The rumour spreads as the post is shared and some form of evidence may be added in the process.
    \item Others begin to challenge its credibility.
    \item A consensus begins to emerge
\end{enumerate}

This can be compared to figure \ref{fig:conversation}, where \textit{user 0} starts a rumour and several other users replies, some challenging its credibility by either querying the rumour(\textit{user 2}) or denying it(\textit{user 4}). Had the example been bigger we might also have seen other people actually re-posting the initial post, some supporting it with URLs to the incident, and after some time a general consensus could possible be inferred from the messages.

\subsection{Annotation}
\label{annotation}
When data is gathered and features are extracted, the question is then which kind of labels one should use for annotation. 
One annotation scheme that is popular is SDQC which labels a text as either \textbf{s}upporting, \textbf{d}enying, \textbf{q}uerying or \textbf{c}ommenting in regards to some rumour. This is discussed in section \ref{sdqc}, followed up by a comparison to topic classification in section \ref{topic}, which tries to label a text to be within some predefined category. \\

Manually annotating data does however come with some challenges. It is time consuming to have experts and individuals annotate the data manually and the annotations could be influenced by the individuals' personal bias. Different annotators might have different views of which labels are appropriate for some microblog post. One example is \cite{AnnotateItalian_TW-BS}, where 8 annotators manually annotate over 8000 tweets. Each tweet is annotated twice by different annotators, and there are disagreements on more than 2000 of the tweets. To mitigate the disagreements from personal bias, a crowd sourcing platform is utilised to give another set of annotations\footnote{\url{https://en.wikipedia.org/wiki/Figure_Eight_Inc.} Visited 09-12-2018}.

Not only is it important which labels are used, but also \textit{what} data is being annotated. Twitter is a popular platform for gathering data. It facilitates an easy way to gather large amounts of text data which can circumvent controversial debates or events. Using public datasets in research should help enabling others to verify and improve on prior research. 

While Twitter is a great platform for gathering data, it is not the only source of data out there and this must be kept in mind. If data from Twitter is primarily used, one could think that it might skew models and systems to be optimised for text written in the context of that particular social media platform and might be less useful elsewhere. This is further discussed in section \ref{twitter_conversations}.

\subsubsection{Labels and SDQC}
\label{sdqc}
The idea of using the SDQC labels stems from an article, which experimentally analyses tweets sent during the August 2011 riots in England with a ``computationally assisted methodology"\cite{procter13}. They develop a code frame for annotating rumour tweets with 6 labels: claim with and without evidence, counterclaim with and without evidence, appeal for more information, and comment. This framework is extended and used in a more recent study, in which they develop a methodology that enables to collect, identify and annotate rumour data from Twitter\cite{zubiaga16}. They assume two types of tweets, namely the source tweet that initiates the rumour and the response tweets that respond to it(See also figure \ref{fig:conversation}). They categorise their labels in three main dimensions, which express a mode of interaction; support/response type, certainty, and evidentiality. 

\textit{Support/response type} is dependent on the tweet type, where a source tweet can be labelled as supporting, denying or under-specified in regards to the content of the statement. If it is a response tweet, it can be labelled as agreed, disagreed, appeal for more information, and comment. These labels corresponds to the codes using in the formerly mentioned paper, \cite{procter13}. In addition to their work, however, \cite{zubiaga16} also consider response types for nested responses, i.e. tweets not directly responding to the source tweet. 

The \textit{certainty} measures the degree of confidence expressed by the author of a tweet when posting a statement in the context of a rumour. The values for the dimension include; certain, somewhat certain, and uncertain. Finally \textit{evidentiality} determines the type of evidence, if any, provided directly in relation to the rumour being discussed. This includes seven labels, where attachments and quotation are examples. \\


The methodology described above is the general approach for the articles investigated in this paper as most of them work with data following the format described in section \ref{data_structure}. An important take-away is the observation that nested posts/replies play a big role for the propagation of rumours.


\subsubsection{A related annotation scheme: Topic classification}
\label{topic}
SDQC seems to be a fair annotation scheme, as the labels divide the classes into very general opinion categories, supposedly making it very suitable for stance classification. In comparison we could look at a different approach and investigate another annotation scheme. One such example could be \textit{topic classification}. 

Topic classification is somewhat similar to stance classification, but differs in its objective, and thus its annotation scheme. Where the latter deals with classifying opinions(stance) from text, the former deals with classifying specific topics from the content of text. This approach is used in \cite{gimenez17} to analyse tweets in regards to the Spanish election in 2015. They introduce five categories for topic labelling: (1) political issues, (2) policy issues, (3) personal issues, (4) campaign issues, and (5) other issues. 

In regards to SDQC, where the labels are rather general and can be used in any stance classification task, this annotation scheme is rather context specific. 
They conclude that the task was complicated, in particular when the topics were similar. One can indeed imagine that the tweet data would contain text with generally more than one of the topics included, making it difficult to annotate it with only one category. With SDQC, you would typically not see a person both support \textit{and} deny some claim. Thus stance classification is more forgiving in comparison to topic classification when it comes to the annotation scheme.

\subsection{Twitter conversations as a social phenomenon}
\label{twitter_conversations}
The methodology behind the SDQC annotation scheme is analysed in \cite{tolmie18}, where they compare the sociological aspects of conversations and compare them to that of microblogging, and in particular Twitter. They conclude that microblogging cannot be treated as a face-to-face conversation due to various factors, including the asynchronous nature of the technology and limits in messages. They investigate microblogging as a turn-taking system, in which one person initiates a message(and potentially a rumour), from which users take turn in responding to. One interesting observation in this regard is that the flow in face-to-face conversations allows for naturally ``self-selecting" next speakers, whereas there are no turn order in microblogging because of the temporal gaps. They find out that rumours unfold across multiple turns and that one needs to examine the organisational characteristics of how specific phenomena unfold in social interaction in order to understand how they work as social phenomena. This means that focusing on one tweet in isolation is very limiting in regards to the information that can be extracted in a social context. The annotation schema is then based on the following observations in relation to Twitter, where 4 and 5 are specifically related to the production of rumours:
\begin{enumerate}
    \itemsep0em 
    \item Tweets are sequentially ordered
    \item Exchanges involve topic management
    \item Important accountability mechanisms are in play
    \item Agreement and disagreement
    \item How tweets are rendered trustworthy through production or evidence
\end{enumerate}
More specifically 2 and 3 relate to the task of labelling source tweets as either aligning or refuting a news event. 1, 2, 3, and 4 relate to the task of labelling whether replies agree or disagree with the source tweet, and finally certainty and evidentiality relate to 5. \\

\subsubsection{Big data on social media}
Related to the subject of Twitter conversations in a social context, \cite{tufekci14} is a research paper on big data on social media, in which methodological and conceptual challenges for this field are studied. Validity and representativeness of social media big data is in focus. Issues in this regard are introduced, in which some of them are of particular interest in the context of stance classification and fake news detection.

One is the ``model organism problem", in which the use of a few specific social media platforms are frequently used to generate data with no consideration of potential bias. It is argued that online platforms such as Twitter raises important questions of representation and visibility because of the difference in behaviour depending on demography and social groups. The point is that we might miss out on important data by making use of the same platforms over and over again.

Another interesting issue is that big data analysis typically relies only on a single social media platform, whereas it is rarely the case that such information is only confined to one source. It is argued that such analysis must take into account that there may be effects which are not visible because relevant information is missing. Thus a request for more research on more than one platform is given in order to understand broader patterns of connectivity.

Finally a point on human interaction on social media platforms argue that human self-awareness needs to be taken into account in big data analysis as humans behave differently when they know they are being observed. Along these lines it is argued that one should take into account that people often are aware of the mechanisms involved in social media communication, and as such can exploit it for their own benefit. This is also related to the concept of \textit{confirmation bias} which is discussed in section \ref{social_and_psycho}. \\

To summarise, even if a social media platform such as Twitter provides easily available data on news event, one should consider the actual data content. It is important to investigate whether the data is representative, whether other platforms can contribute and who the users communicating are and how they behave.

%% file: approaches.tex
\section{Classification approaches}
\label{approaches}
Once data has been gathered and annotated and features extracted, one must decide which approach to use for the actual classification system. This section will provide an overview of different approaches, as well as their results(section \ref{performance_overview}), both for sole stance classification but also applied to fake news detection.  

\subsection{Recurrent Neural Network and Long-Short Term Memory}
\label{rnn_and_lstm}
Recurrent neural network(RNN) systems allow representing arbitrarily sized structured inputs in a fixed-size vector, while paying attention to the structure properties of the input. This makes it quite appropriate for the type of data for stance classification. \cite{kochlina17} implements such a system in SemEval-2017 Task 8 subtask A, in which they implement a Long Short-Term Memory(LSTM) version, feeding in whole conversation branches as input, as opposed to the typical case of using words alone as input. They end up with very nice results, even coming out as the best performing system for the task.

\cite{augenstein16}, as earlier mentioned,  utilises the LSTM model differently. They use the LSTM to facilitate an encoding of target text and source text, using a ``Bidirectional conditional LSTM" approach. That is, input is fed into the network in both directions, and one layer is dependent on another layer's state, thus making it conditional.

\cite{transferLearning} also implements an RNN, but takes an interesting approach by pre-training their model on existing related twitter data utilising the most frequent hash-tag as a label. The lack of domain-labelled data was a challenge in their project and the pre-training of their model was an attempt to tackle this, and yielded good results in comparison to a randomly initialised model. 

\subsection{Support Vector Machine}
Another SemEval-2017 Task 8 paper, which deals with both subtask A and B is \cite{enayet17}. They have focus on the latter(veracity prediction), in which they scored best. They use a linear Support Vector Machine (SVM) approach with BOW features and other manually selected features. The SVM maps data to points in space, and then assign classes to the data depending on the positions of output, as opposed to the probabilistic classification approach typically used in neural networks. 

\subsection{Convolutional Neural Network}
Another approach based on neural networks is that of convolutional neural networks, which works particularly well with spatial data structures.

\cite{chen17} implements a CNN model with leave-one-out(LOO) testing to classify with SDQC. They use varying window sizes, while also training 5 models independently and using majority voting to find results. A similar approach is implemented in \cite{kim14}, which deals with seven different NLP tasks, experimenting with different CNN architectures. It is shown that a CNN with ``static"(non-trained in backpropagation) word embeddings and ``non-static"(trained in backpropagation) word embeddings performs well, whereas a combination of the two, denoted as ``CNN-multichannel" overall performs best. As such it is concluded that unsupervised pre-training of word vectors is an important factor for NLP deep learning, and that a one-layered CNN can perform remarkably well.

\subsection{Crowd stance}
One approach for fake news detection, is to analyse the replies on the source post of the news. Applying stance classification on microblog conversation data enables analysis of the general stance/opinion of the ``crowd"\cite{fakenewschallenge,derczynski17}. If there for example is a lot of negative response to a post, it might show that the crowd is sceptical about the claim/rumour in the source and vice versa. 

\subsection{Hidden Markov Model}
\label{hmm}
One example of a crowd stance implementation is the use of a Hidden Markov Model(HMM) in \cite{dungs18}, which uses stance and tweets' times alone for automatic rumour veracity classification, achieving state-of-the-art results. HMM is well known for temporal pattern recognition, which they utilise as follows: Regard individual stances over a rumour's lifetime as an ordered sequence of observations, and then compare sequence occurrence probabilities for true/false. Their results are obtained using gold stance labels, but they also test it with automatically generated stance labels\cite{aker17}, and observe only a marginal decrease in performance. This shows that their veracity classification system has viable practical applications.

Furthermore they apply their system in a setting for early detection of rumours. That is, they limit the number of tweets to the first 5 and 10 tweets respectively. Surprisingly, even with only 5 tweets, their model still outperforms the baselines, which use all of the tweets. 

\subsection{Tree classification}
\label{tree}
In \cite{aker17} one of the approaches used is a J48 decision tree classifier, which is build over a number of features from earlier work, extended with some ``problem-specific" features. The approach reaches state-of-the-art performance for stance classification, and shows that a simple classifier can work really well. A lot of work went into defining the features for a tweet and shows that the features used to define a tweet are key to having good results.

\subsubsection{Ensemble with CNN}
\label{tree_ensemble}
Another interesting approach is that of the winning team for the Fake News Challenge\cite{fakenewschallenge}, which uses an ensemble of decision trees and a CNN\cite{talos17}. Specifically their model is based on a 50/50 weighted average between gradient-boosted decision trees and a deep CNN. Neither of the models have impressive accuracy, but the 50/50 weighting in the final classification step improves their results. 

\subsection{Multi-Layered Perceptron}
The second best scoring team in the Fake News Challenge\cite{fakenewschallenge}, utilises a Multi-Layered Perceptron(MLP) \cite{athene17} approach with several ReLU layers. The final system ensemble 5 instances of the MLP model and decides the output label by majority voting between the instances. The team scoring third best in the Fake News Challenge also implements an MLP, employing lexical and similarity features with one hidden ReLU layer\cite{riedel2017fnc}. As noted in the paper, the results are quite disappointing since the model is inaccurate at predicting the most interesting classes, ``agree" and ``disagree".



\subsection{Matrix factorization}
In \cite{TriFN} an approach trying to exploit a tri-relationship between publishers, news contents and social engagements is used to do fake news detection. The approach(denoted ``TriFN") utilises non-negative matrix factorisation to generate latent feature vectors for users and news. A comprehensive mathematical model describing the problem as a minimisation problem is described and formally optimised. The results of the TriFN framework outperforms a number of baselines and yields positive results.

\subsection{Bayes classification}
A more mathematical approach than those described so far is classification with Bayes classifiers, as implemented in \cite{qazvinian11}, being a pioneering paper in the area of stance classification and fake news detection. The approach is based on learning a linear function of different Bayes classifiers as high level features to predict whether a user believes a specific rumour or not. Each of the classifiers calculates the likelihood ratio for a given tweet to be under either a positive model or negative model with respect to the given feature(i.e. Bayes classifier). 

\subsection{Performance overview}
\label{performance_overview}
In appendix \ref{app:results_overview} a brief overview of the used method, dataset and results for the papers presented in this section is shown in table \ref{tab:performance_overview}. Note that this is not meant to be a direct comparison of the approaches taken in the papers, as some of the papers use different metrics, degree of classification and datasets and a comparison would as such not give much value. 

The results for the Fake News Challenge systems are using a custom metric, making it difficult to compare with other systems. However, \cite{hanselowski18} have reproduced the results and reports their $F_1$ score, which are used here as well. Interestingly the teams' rankings change when comparing $F_1$ scores. Note also that the lower part of table \ref{tab:performance_overview}(divided by double-lines) reports results for binary classification for the fake news detection task, whereas the upper part is for stance classification.

%% file: proposal.tex
\section{Proposal of system implementation approach}
\label{proposal}
In this section we propose a concrete system implementation for the thesis project as a result of the research in this paper. The objective is to build a stance classification system for Danish over social media data, and apply it to rumour resolution and possible to detect fake news. As such we need to gather data, annotate it, build a classification system and deploy it\footnote{By deploying it we mean to make it publicly available to use ``out-of-the-box" on GitHub}. Furthermore we include a tentative thesis plan as a feasibility check and guideline for the project.

\subsection{Data gathering and annotation}
\label{data_proposal}
For the data gathering and annotation phase we propose to use the social media platform, Reddit, and in particular the official Danish sub-reddit on \url{https://www.reddit.com/r/Denmark/}. Reddit has an open-source API allowing to use their data for non-commercial use\footnote{\url{https://www.reddit.com/wiki/api} Visited 04-12-2018}. Clearly Twitter would be an obvious candidate to choose as is clear from the research presented in this paper, but we have spent some time on exploring Danish news on this platform, which is not really present. Facebook would be another good candidate, but its data is not publicly available. 

For annotation we propose to use the SDQC approach, that is, four labels indicating: support, denying, querying, and commenting. \\

Alternatively, if we do not succeed in finding a proper microblog platform, we can use the same approach as the task in the Fake News Challenge\cite{fakenewschallenge}, i.e. performing stance classification based on a headline and a body text for a news event.

\subsection{Classification and detection system}
We propose to build a stance classifier model with a decision tree approach, which is covered in section \ref{tree}, as it has proved to be a simple yet effective approach for stance classification.

Further, the approach described in section \ref{hmm} is quite interesting, achieving very nice results, implementing a HMM for veracity detection using few and simple features, including stance and tweets' times. We propose to use the same approach for fake news detection based on historical events.

In the case where we do not have data in the form of a microblog, but as news articles(see section \ref{data_proposal} above), we propose to use an ensemble approach as the one introduced in section \ref{tree_ensemble}, combing tree classification with deep learning methods.

\subsection{Technology}
We also propose a framework and programming environment to work with for the project. Python is a popular programming language for data analysis, including ML and NLP, because of its plethora of libraries. A HMM model could be implemented using \texttt{hmmlearn}\footnote{\url{https://github.com/hmmlearn/hmmlearn}}, a neural network could be implemented using PyTorch\footnote{\url{https://pytorch.org/}}, and a decision tree with the scikit-learn library\footnote{\url{https://scikit-learn.org/}}. Python also features a rich library for data pre-processing, NLKT\footnote{\url{https://www.nltk.org/}}, which among other things does automatic word tokenization. Apart from the useful libraries, Python is a high-level language allowing for a code-first approach, i.e. fast prototyping.
           
\subsection{Thesis plan}
The thesis project is carried out February through June 2019, with deadline for hand-in June 3\textsuperscript{rd} at 14.00 o'clock. Thus we have 17 weeks to implement the system and write the thesis paper. A high-level thesis plan is sketched in table \ref{tab:thesis_plan}, appendix \ref{app:thesis_plan}.

The plan comprises of work items for the implementation of the system and thesis sections divided into the weeks we intend to carry them out. The first month the focus will be on data gathering and annotation. Simultaneously and the following month we will build a prototype for an early evaluation of the data. The third month we will tune the system, run experiments and change the parameters accordingly, allowing us to test and evaluate the system. Then we will deploy it, making it publicly available.

The plan also consists of a week by week overview of the sections we will focus on in the thesis paper. First off the general structure of the thesis as well as data gathering will be in focus. In the following weeks the annotation, baseline and dataset will be covered. After this the choice of technology, data analysis and the prototype systems will be discussed. Then the parameter space, optimal parameters and results will be reported after running experiments. Finally the results will be analysed and discussed before concluding the thesis.

%% file: conclusion.tex
\section{Conclusion}
\label{conclusion}
The objective for this research paper has been to survey stance classification and fake news detection. The task of classifying the opinion of a crowd towards a rumour has been explored by many approaches within the last 10 years, resulting in very useful findings. One particularly interesting use of stance classification is to use it for assessing whether some news event is true or false. One challenge in this regard is the social and psychological aspects occurring in microblogs, where polarised opinions take effect because of filter bubbles and the echo chamber effect. Another challenge for analysing data from microblogs, such as Twitter, is the model organism problem, which is the prevalent issue of representation and visibility when continuously using the same platforms in stance classification. 

We have further investigated the process of data gathering and annotation, and how imbalanced data can have significant impact of the results obtained in stance classification. In particular feature engineering seem to be of great importance, choosing representative information to extract that will do great on the test data at hand while still being general-purpose oriented. 

Different methods for stance classification and fake news detection have been explored, but because of different data and metrics it has been difficult to directly compare their results. However, one particular approach is very relevant and interesting for the thesis project, which is the use of a HMM in analysing rumours in microblog data, achieving very promising results. Furthermore the use of a decision tree model for stance classification appear to be a good choice. These are also the approaches we propose to use in the thesis project, where we intend to gather a dataset in the Danish language, annotate it, build the classifier/detection system and deploy it. In conclusion the findings in this research paper will most likely prove very useful as background knowledge in the coming thesis project.

%% file: appendix.tex
\onecolumn
\section{Results overview}
\label{app:results_overview}
\begin{table}[h]
    \centering
    \begin{tabular}{|l c c l|}
        \hline
        \textit{Approach} & \textit{Acc} & $F_1$ & \textit{Dataset} \\
        \hline
        \hline
        Transfer Learning RNN & - & 67.8 & \cite{mohammad16} \\
        \hline
        Bidirectional LSTM & - & 58.3 & \cite{mohammad16} \\
        \hline
        Branch LSTM & - & 43.4 & \cite{derczynski17} \\
        \hline
        CNN & - & 53.6 & \cite{derczynski17} \\
        \hline
        SVM & 53.0 & - & \cite{derczynski17} \\
        \hline
        J48 & 79.02 & - & \cite{derczynski17} \\
        \hline
        MLP & - & 60.4 & \cite{fakenewschallenge} \\
        \hline
        MLP & - & 58.3 & \cite{fakenewschallenge} \\
        \hline
        CNN and Tree ensemble & - & 58.2 & \cite{fakenewschallenge} \\
        \hline
        \hline
        Bayes & 94.1 & 92.5 & \cite{qazvinian11} \\
        \hline
        Hidden Markov Models & - & 80.4 & \cite{zubiaga16} \\
        \hline
        TriFN & - & 87.0 & \cite[BuzzFeed]{TriFN} \\
        \hline
        TriFN & - & 88.0 & \cite[Politifact]{TriFN} \\
        \hline
    \end{tabular}
    \caption{Overview of performance results for the different approaches for stance classification(top) and fake news detection(bottom)}
    \label{tab:performance_overview}
\end{table}

\clearpage
\section{Thesis plan}
\label{app:thesis_plan}
\begin{table}[h!]
    \centering
    \begin{tabular}{r p{4cm} p{4cm} p{4cm}}
        \textit{Week} & \textit{Work item} & \textit{Thesis} & \textit{Milestones} \\
        \hline
        6 &  Data gathering & General structure and data gathering & \\
        7 &  Data annotation and \newline prototype & Annotation & \\
        8 &  Data annotation and \newline prototype & Baseline & \\
        9 &  Data annotation and \newline prototype & Dataset & \\
        \hline
        10 & Prototype testing & Technology & Working prototype and gathered dataset \\
        11 & Data evaluation & Data analysis and \newline statistics & \\
        12 & Finalize prototype & Describe prototype \newline system & \\
        13 & Finalize prototype & Describe prototype\newline system & \\
        \hline
        14 & Tune system\newline parameters & Parameter space & Intermediate results and finished prototype\\
        15 & Tune system\newline parameters & Optimal parameters & \\
        16 & Test & Experiment results &\\
        17 & Test & Draft for final version &\\
        \hline
        18 & Evaluation of results & Result and error\newline analysis & Draft for thesis and results gathered\\
        19 & Evaluation of results & Result and error\newline analysis & \\
        20 & System revision & Discussion & \\
        21 & Conclude & Conclude, abstract & \\
        22 & Deploy system & Proof read & \\
        \hline
    \end{tabular}
    \caption{Thesis plan}
    \label{tab:thesis_plan}
\end{table}